\documentclass[electronic]{vgtc}             

\ifpdf
  \pdfoutput=1\relax                   
  \usepackage{graphicx}                
  \DeclareGraphicsExtensions{.pdf,.png,.jpg,.jpeg} 
\else
  \usepackage{graphicx}                
  \DeclareGraphicsExtensions{.eps}     
\fi%

\graphicspath{{figures/}{pictures/}{images/}{./}} 

\usepackage{microtype}                 
\PassOptionsToPackage{warn}{textcomp}  
\usepackage{textcomp}                  
\usepackage{mathptmx}                  
\usepackage{times}                     
\usepackage{cite}                      
\usepackage{tabu}                      
\usepackage{booktabs}                  

\usepackage{stmaryrd}
\usepackage[mathscr]{euscript}
\usepackage{tabularx}
\usepackage{amsmath}
\usepackage{amssymb}

\usepackage{mathtools}
\newcommand{\defeq}{\mathrel{\triangleq}}
\newcommand{\retr}{\mathrm{retr}}
\newcommand{\rel}{\mathrm{rel}}
\usepackage{fancybox}

\usepackage{xspace}
\newcommand{\methodname}[1]{\textsf{#1}\xspace}
\newcommand{\ming}{\methodname{MING}}
\newcommand{\cviz}{\methodname{CheckViz}}
\newcommand{\ddhds}{\methodname{DD-HDS}}

\renewcommand{\ovalbox}[1]{\methodname{#1}}

\onlineid{0}

\vgtccategory{Research}

\vgtcinsertpkg

\title{Interpreting Distortions in Dimensionality Reduction by Superimposing Neighbourhood Graphs}

\author{Benoît Colange\textsuperscript{1,2}\thanks{benoit.colange@cea.fr} 
\and Laurent Vuillon\textsuperscript{2} %
\and Sylvain Lespinats\textsuperscript{1} %
\and Denys Dutykh\textsuperscript{2}}
\affiliation{\scriptsize \textsuperscript{1}Univ. Grenoble Alpes, INES, F-73375, Le Bourget du Lac, France \\ \textsuperscript{2}Univ. Grenoble Alpes, Univ. Savoie Mont Blanc, CNRS, LAMA, 73000 Chambéry, France}

\abstract{
	To perform visual data exploration, many dimensionality reduction methods have been developed. 
   These tools allow data analysts to represent multidimensional data in a 2D or 3D space, 
   while preserving as much relevant information as possible. 
   Yet, they cannot preserve all structures simultaneously 
   and they induce some unavoidable distortions. 
   Hence, many criteria have been introduced to evaluate a map's overall quality, 
   mostly based on the preservation of neighbourhoods. 
   Such global indicators are currently used to compare several maps, 
   which helps to choose the most appropriate mapping method and its hyperparameters. 
   However, those aggregated indicators tend to hide the local repartition of distortions. 
   Thereby, they need to be supplemented by local evaluation to ensure correct interpretation of maps. 
   
   In this paper, we describe a new method, called \ming, for ``Map Interpretation using Neighbourhood Graphs''. 
   It offers a graphical interpretation of pairs of map quality indicators, 
   as well as local evaluation of the distortions. 
   This is done by displaying on the map the nearest neighbours graphs computed in the data space and in the embedding. 
   Shared and unshared edges exhibit reliable and unreliable neighbourhood information conveyed by the mapping. 
   By this mean, analysts may determine whether proximity (or remoteness) of points on the map faithfully represents similarity (or dissimilarity) of original data, within the meaning of a chosen map quality criteria. 
   We apply this approach to two pairs of widespread indicators: precision/recall and trustworthiness/continuity, chosen for their wide use in the community, which will allow an easy handling by users.} 

\CCScatlist{
  \CCScatTwelve{Evaluation}{Qualitative Evaluation}{}{};
  \CCScatTwelve{Non-Spatial Data and Techniques}{Dimensionality Reduction}{}{};
  \CCScatTwelve{Visual Analysis and Knowledge Discovery}{Visual Knowledge Discovery}{}{}
}

\begin{document}


\firstsection{Introduction}

\maketitle

Dimensionality reduction aims at representing $N$ points $\xi_{\,i}$, lying in a high dimensional metric data space, 
by a set of corresponding points $x_i=\Phi(\xi_{\,i})$ in an embedding space of lower dimensionality (with $\Phi$ a mapping). 
To obtain this mapping, many techniques have been developed, including \methodname{classical MDS} \cite{torgerson_multidimensional_1952}, \methodname{Isomap} \cite{tenenbaum_global_2000}, 
\ddhds \cite{lespinats_dd-hds:_2007} and \methodname{tSNE} \cite{maaten_visualizing_2008}. 
In order to perform visual exploration of data, the embedding dimensionality is often chosen to be 2 or 3. 
This task mainly relies on the hypothesis that two similar data points must be embedded close from each other, 
whereas dissimilar points must be represented far apart.
However, the mapping process often introduces distortions, which invalidate this premise. 
As a consequence, the proximity or remoteness of points on a map alone is not sufficient to infer information concerning structures (\textit{e.g.} underlying manifold, clusters or outliers \cite{lee_nonlinear_2007, sacha_visual_2017}) in the original data. 
The considered visualization must be enhanced with local evaluation, locating precisely the distortions measured through global indicators.

In the present article, we introduce a new local evaluation method called \ming, for ``Map Interpretation using Neighbourhood Graphs''. 
This approach adapts global rank-based indicators (\textit{e.g.} precision and recall \cite{venna_information_2010} or trustworthiness and continuity \cite{venna_neighborhood_2001-1}), designed for assessing overall performances of dimensionality reduction, to the needs of local evaluation. 
It displays on the map neighbourhood graphs, which link each point to its $\kappa$ nearest neighbours, 
in order to exhibit distortions. 
On these graphs (considered successively in the embedding and data space), edges corresponding to reliable or unreliable neighbours are then distinguished using different colours, 
colour range being chosen so as to offer a local representation of the selected global indicators. 

\subsection{Notations}
$\Delta_{\:ij}$ is the distance in the data space between two points $\xi_{\,i}$ and $\xi_j$, 
and $\rho_{\,ij}$ is the rank of $\xi_j$ in the neighbourhood of $\xi_{\,i}$, 
meaning that $\xi_j$ is the $\rho_{\,ij}^{\,\mathrm{th}}$ nearest neighbour of $\xi_{\,i}$, with $\rho_{\,ii}=0$ by convention. 
Conversely to distances, ranks are not necessarily symmetric ($\rho_{\,ij}$ may be different from $\rho_{ji}$). 
Distances $D_{\,ij}$ and ranks $r_{\,ij}$  may be defined equivalently in the embedding space. 
In the scope of this article, the $\kappa$ neighbourhood of the $i^\mathrm{\,th}$ point in a given space is defined as the set of its $\kappa$ nearest neighbours. It is denoted as a set of those neighbours' indices: $\nu_i(\kappa)$ in the data space and $n_i(\kappa)$ in the embedding space.

\subsection{Distortions as corruption of neighbourhood graphs}
\begin{figure}[htb]
  \centering
  \includegraphics[width=.8\columnwidth]{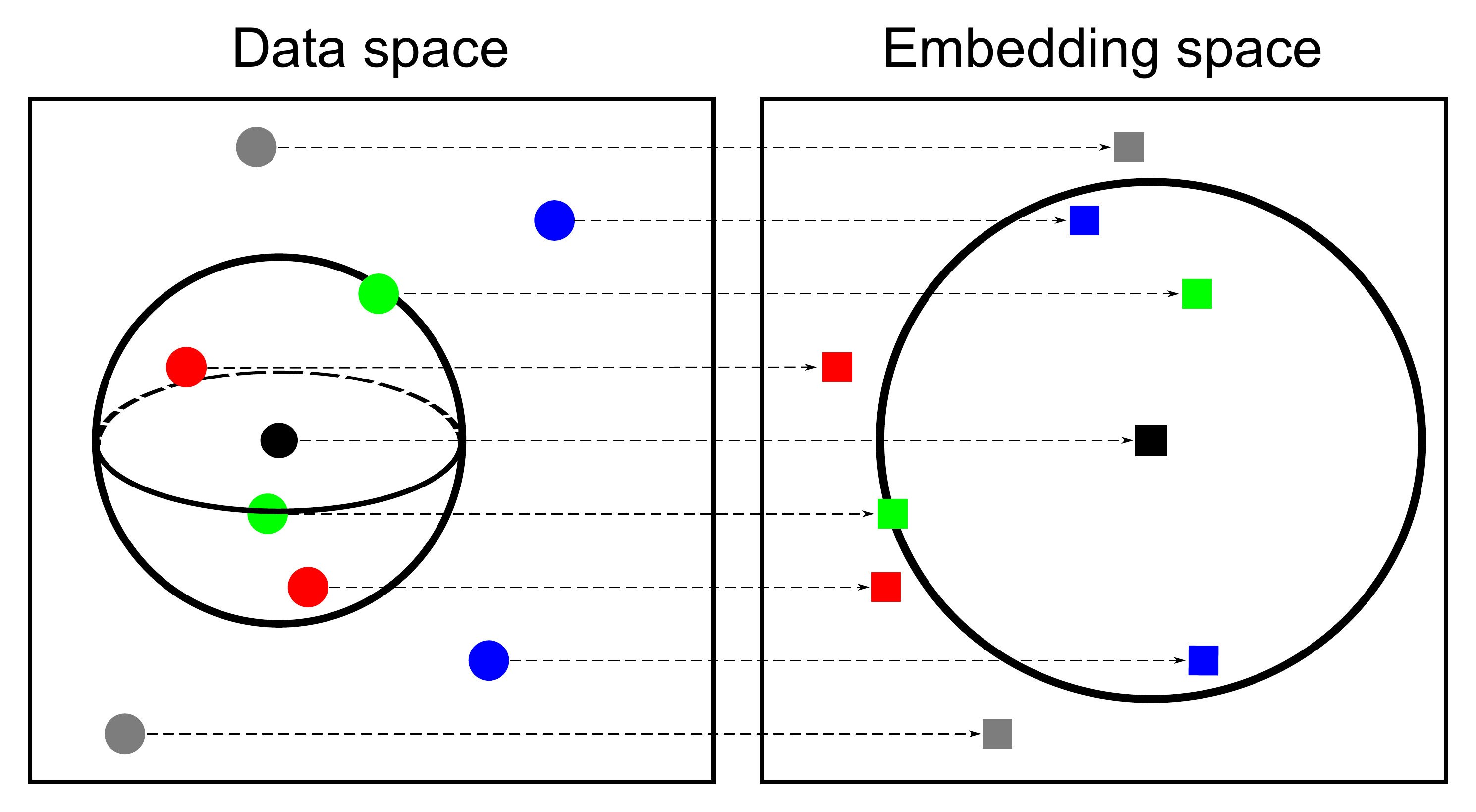}
  \caption{\label{fig:DistortionTypes}
  Preservation and distortions of $4$-neighbourhood relations $(i,j)$ for a point $i$ (black), with reliable (green), false (blue), missed (red) and non-existent (grey) neighbours, and $4$-neighbourhoods $\nu_i(4)$ and $n_i(4)$ materialized by the black sphere and circle.
           }
\end{figure}

Herein, we define distortions as a defect of preservation of \textit{neighbourhood relations}. At a scale $\kappa$, the relation $(i,j)$ is said to exist in a given space if the point $j$ belongs to the $\kappa$ neighbourhood of point $i$ in that space. 
Those relations may easily be interpreted as edges of neighbourhood graphs. 
For a specific mapping, the neighbourhood relation $(i,j)$ may be qualified as reliable if it exists in both spaces, false if it exists only in the map, missed if it exists only in the data space, or non-existent if it exists in neither of those. 
Figure~\ref{fig:DistortionTypes} sums up these various cases, $\kappa$ neighbourhoods being represented as balls centred on $i$ and whose radius is the distance to the $\kappa^{\,\mathrm{th}}$ neighbour.

\section{Related work}
Several indicators are currently used in order to estimate a map's quality. 
They are generally based on the comparison of:
\begin{itemize}
\setlength\itemsep{-0.5em}
\item distances $\Delta_{\:ij}$  and $D_{\,ij}$, used in stress functions of many dimensionality reduction methods \cite{torgerson_multidimensional_1952, sammon_nonlinear_1969, demartines_curvilinear_1997}
\item ranks $\rho_{\,ij}$ and $r_{\,ij}$, used in several rank-based quality criteria \cite{lee_quality_2009}. Due to their robustness to the phenomenon of norm concentration occurring in high dimensionality spaces, 
ranks tend to be more considered.
\end{itemize}
Global and local evaluation consider the distortions at three levels of aggregation. 
\begin{enumerate}
\setlength\itemsep{-0.5em}
\item Map-wise aggregation allows to produce global indicators for evaluating mappings (see Section \ref{sec:GlobalEval}). Those are computed by averaging point-wise values.
\item Point-wise aggregation of pairwise penalizations is often displayed by local evaluation methods, as mentioned in Section \ref{sec:LocalEval}.
\item Pairwise penalizations are constructed for each neighbourhood relation $(i,j)$, assessing its level of distortion. Though such penalizations are commonly established, previous works tend to either aggregate them, losing the pairwise information, or to display them for only one reference point at a time. In Section \ref{sec:MING}, we introduce \ming to display the pairwise penalizations on neighbourhood graphs.
\end{enumerate}

\subsection{Global evaluation}\label{sec:GlobalEval}
In order to compare different mappings, quantitative measures have been developed, producing scalar indicators. Many of these go by pairs $\bigl(\mathscr{F},\,\mathscr{M}\bigr)$, with $\mathscr{F}$ penalizing false neighbours and $\mathscr{M}$ penalizing missed neighbours.
We focus here on widely used rank based criteria: precision $\mathscr{P}$ and recall $\mathscr{R}$ \cite{venna_information_2010},
and trustworthiness $\mathscr{T}$ and continuity $\mathscr{C}$ \cite{venna_neighborhood_2001-1}. 
However, the adopted framework, and thus our proposed method, may easily be extended to other rank based criteria, such as \methodname{Mean Relative Rank Error} (\methodname{MRRE}) \cite{lee_nonlinear_2007}, or, with minor changes, to distance based criteria (\textit{e.g.} \methodname{Curvilinear Component Analysis} (\methodname{CCA}) \cite{demartines_curvilinear_1997} and \methodname{Sammon's mapping} \cite{sammon_nonlinear_1969} stress functions).
At a given scale $\kappa$, the considered rank-based criteria are computed as detailed below for each pair of indicators $(\mathscr{F},\mathscr{M})$.

Pairwise penalizations $\mathscr{F}_{ij}(\kappa)$ and $\mathscr{M}_{ij}(\kappa)$ are defined for all penalized neighbourhood relations $(i,j)$, accounting for their distortions. 
Those penalized relations are restricted for each point $i$ to the set of $\kappa$ neighbourhood relations existing at least in the embedding space $\{(i,j)\mid j\in n_i(\kappa)\}$ for false neighbours indicator $\mathscr{F}$, and to the set of neighbourhood relations existing at least in the data space $\{(i,j)\mid j\in \nu_i(\kappa)\}$ for missed neighbours indicator $\mathscr{M}$. 
Table~\ref{table:PairwisePenalizations} sums up the values of these pairwise penalizations for the two selected pairs of indicators. 
With these penalizations, we can distinguish reliable relations with null weights from distorted relations (false or missed) with strictly positive weights. 
To calculate global indicators, based on this pairwise information, point-wise (Equations~\ref{eqn:PointwiseAggregFalse} and \ref{eqn:PointwiseAggregMissed}) and map-wise (Equation~\ref{eqn:MapwiseAggreg}) aggregations are then performed successively: 
\begin{equation}
\label{eqn:PointwiseAggregFalse}
\mathscr{F}_i(\kappa)\defeq \sum_{j \in n_i(\kappa)}  \mathscr{F}_{ij}(\kappa),
\end{equation}
\begin{equation}
\label{eqn:PointwiseAggregMissed}
\mathscr{M}_i(\kappa)\defeq \sum_{j \in \nu_i(\kappa)}\mathscr{M}_{ij}(\kappa),
\end{equation}
\begin{equation}
\label{eqn:MapwiseAggreg}
\mathscr{I}(\kappa)\defeq 1-\frac{1}{N}\sum_{i=1}^N \frac{\mathscr{I}_{\,i}(\kappa)}{C_\mathscr{FM}(\kappa)} \mbox{ for }\mathscr{I}\in\{\mathscr{F},\mathscr{M}\},
\end{equation}
where $C_\mathscr{FM}(\kappa)$ is the upper bound of the point-wise values, 
so that criteria go from $0$ for worst mapping (reversed ranks) to $1$ for an ideal mapping.
\newcolumntype{Y}{>{\centering\arraybackslash}X}
\noindent
\begin{table}[htb]
\caption{\label{table:PairwisePenalizations} Pairwise penalizations and sets of considered neighbours indices for precision/recall and trustworthiness/continuity.}
\scriptsize
\begin{tabularx}{\columnwidth}{| c | Y | Y |}
\hline
$\bigl(\mathscr{F},\,\mathscr{M}\bigr)$ & 
$\mathscr{F}_{ij}(\kappa) \text{ for } j \in n_i(\kappa)$ &
$\mathscr{M}_{ij}(\kappa) \text{ for } j \in \nu_i(\kappa)$\\
\hline \hline
$\bigl(\mathscr{P},\,\mathscr{R}\bigr)$ &
$\left\{\begin{array}{r} 1 \text{ if } j \notin \nu_i(\kappa) \\ 0 \text{ if } j \in \nu_i(\kappa) \end{array}\right.$ & 
$\left\{ \begin{array}{r} 1 \text{ if } j\notin n_i(\kappa) \\ 0 \text{ if } j\in n_i(\kappa) \end{array}\right.$\\
\hline
$\bigl(\mathscr{T},\,\mathscr{C}\bigr)$ & 
$\left\{\begin{array}{c} \rho_{ij}-\kappa \text{ if } j \notin \nu_i(\kappa) \\ 0 \text{ if } j \in \nu_i(\kappa) \end{array}\right.$ & 
$\left\{\begin{array}{c} r_{ij}-\kappa \text{ if } j\notin n_i(\kappa) \\ 0 \text{ if } j\in n_i(\kappa) \end{array}\right.$\\
\hline
\end{tabularx}
\end{table}

\subsection{Local Evaluation}\label{sec:LocalEval}
Many existing evaluation methods tend to display point-wise estimations of local distortions \cite{seifert_stress_2010, martins_visual_2014, stahnke_probing_2016}. Among them, \cviz \cite{lespinats_checkviz:_2011} provides simultaneously the values of $\mathscr{F}_i$ and $\mathscr{M}_{\,i}$ with a bi-dimensional colour scheme interpolated onto the background, distinguishing areas with false/missed neighbours, or both. Yet \cviz fails to show which pairwise relations suffer from these distortions.

Other approaches focus on a reference point to present its true neighbourhood relations, by colouring the background (\textit{e.g.} \methodname{Proxilens} \cite{aupetit_visualizing_2007, heulot_proxilens:_2013}) or correcting distances \cite{stahnke_probing_2016}. These unveil reliable relations and distortions but for only one point $i$ at a time. Martins et al. \cite{martins_visual_2014} also proposed to focus on subsets of reference points for representing missed neighbours.

\section{Map Interpretation using Neighbourhood Graphs (\ming)}\label{sec:MING}
In the present section, we describe a local evaluation method that:
\begin{itemize}
\setlength\itemsep{-0.5em}
\item is consistent with state of the art global rank based indicators,
\item shows the reliability of $\kappa$ neighbourhood relations between pairs of points (without any aggregation),
\item can be directly applied to data visualization in a 3D embedding space, while partially alleviating depth ambiguity and apparent distances distortions that occur in 3D scatter plots \cite{sedlmair_empirical_2013}.
\end{itemize}
In this approach, we represent the penalized neighbourhood relations as edges of two directed graphs, each associated to one of the indicators. 
Using the interpretation of data visualization as an information retrieval task \cite{venna_information_2010}, the graph associated to the false neighbours indicator $\mathscr{F}$ is called the retrieval graph, defined as follows:
\begin{equation}
\label{eqn:RetrGraph}
G_{\mathscr{F}}^{\retr}(\kappa)\defeq\bigl(V,E^{\retr}(\kappa),W_{\mathscr{F}}^{\retr}(\kappa)\bigr).
\end{equation}
In Equation~\ref{eqn:RetrGraph}, $V$ are the vertices of the graph (corresponding to the points), $E^{\retr}(\kappa)\defeq\bigl\{(i,j) \in V^2 \, \vert \, j \in n_i(\kappa)\bigr\}$ the edges corresponding to penalized neighbourhood relations for $\mathscr{F}$, \textit{i.e.} relations existing at least in the embedding space (reliable and false), and $W_{\mathscr{F}}^{\retr}(\kappa)\defeq\{\mathscr{F}_{ij}(\kappa)\mid (i,j)\in E^{\retr}(\kappa)\}$ the weights associated with these edges and corresponding to the pairwise penalizations defined for the indicator. 

In the same way, the graph associated to the missed neighbours indicator $\mathscr{M}$ is called the relevance graph and may be defined by:
\begin{equation}
\label{eqn:RelGraph}
G_{\mathscr{M}}^{\rel}(\kappa)\defeq\bigl(V,E^{\rel}(\kappa),W_{\mathscr{M}}^{\rel}(\kappa)\bigr),
\end{equation} 
with $E^{\rel}(\kappa) \defeq \bigl\{(i,j) \in V^2 \, \vert \, j \in \nu_i(\kappa)\bigr\}$ the penalized edges for $\mathscr{M}$, \textit{i.e.} relations existing at least in the data space (reliable and missed), and $W_{\mathscr{M},\, ij}^{\rel}(\kappa)\defeq\mathscr{M}_{\,ij}(\kappa)$ their associated weights.\\

The graphs are then visualized by successively superimposing them on the map (namely, positioning vertices $i\in V$ using coordinates of points in the embedding). Distortions introduced by the mapping  are shown by colouring the edges depending on their weights. 
In order to stress the differences between the two graphs, the two kinds of penalizations are colour-coded differently, using \textit{GnBu} for the retrieval graph and \textit{OrRd} for the relevance graph (those two colour maps from \methodname{ColorBrewer} \cite{brewer_colorbrewer} being colour-blind safe). Thus, reliable edges are displayed as white, while false neighbours are shown in shades of green and blue, and missed neighbours in shades of orange and red. 

Neighbourhood relations being asymmetrical, directed edges $(i,j)$ and $(j,i)$ are different. Thus the line connecting points $i$ and $j$ on the map is split into two segments coloured independently. The colour closer from a point $i$ corresponds to the weight of the edge $(i,j)$ (in the direction from $i$ to $j$), and equivalently, the color closer from point $j$ shows the weight in the other direction $(i,j)$. 
For relations existing only in one direction, as is common for outliers, the second part of the segment, associated with the non-existent direction is materialized by a dotted line. 
The best illustration of this may be found on Figure~\ref{fig:MingTrustCont}.

\section{Experiments}
\subsection{Considered datasets}
To illustrate our method, we consider two datasets. 
The first consists in 200 randomly selected samples from the handwritten digits dataset \cite{Dua:2017} ($8\times8$ grey level images, seen as data points in a space of $64$ dimensions). 
These are embedded in a 2D space using \ddhds \cite{lespinats_dd-hds:_2007} (as shown in Figures~\ref{fig:CheckViz}, \ref{fig:MingPrecRec}, \ref{fig:MingTrustCont} and \ref{fig:digits_edge_bundling}). 
This dataset has a high intrinsic dimensionality, so that the mapping process has to introduce many distortions. 
The second is the coil-20 dataset, which consists in $128\times128$ images of $20$ different objects continuously rotated around an axis. 
Consequently, its intrinsic dimensionality is intuitively very low, since the generation of images for one object has only one degree of freedom (which is the rotation angle). Thus, its embedding, performed using \methodname{t-SNE}, may be relatively good.

One key advantage of choosing these images datasets, is that we may use the actual images as markers for points on the map, allowing the reader to assess visually whether two items are similar or not in the data space (within the meaning of Euclidean distances). Note that for a better rendering of these images, the coil-20 data set has been sampled by removing one in two elements (artificially leading to a bigger angular step). 
Finally, the use of circular backgrounds, in the presented maps, intends to remind viewers that axes obtained by MultiDimensional Scaling have no directly accessible meaning \cite{degret_circular_2018}. Indeed, such a map may be translated, rotated or symmetrized without any effect on the information that it conveys. 

\subsection{Choice of \texorpdfstring{$\kappa$}{TEXT}}
In \ming, as in \cviz, the user remains free to set the parameter $\kappa$, 
depending on the number of neighbours they judge relevant. 
Several values may also be used successively during users interaction with the representation, in order to get information for different scales, and on the hierarchical organization of data. 
However, with too high values of $\kappa$, \ming figures tends to be less readable due to visual clutter, especially for the relevance graph, which may be considerably entangled. 
A similar phenomenon occurs for severely distorted mappings, which provide very few reliable neighbourhood relations. 

For the digits dataset, a setting of $\kappa=10$ has been selected, so that the majority of edges remains intra-classes. 
For the coil-20 data set however, points form continuums, so that only a few closest neighbours are sufficient to illustrate the quality of the mapping. Thus, we used a value of $\kappa=4$ for this dataset.

\subsection{\ming interpretation guidelines}
\begin{figure}[htb]
  \centering
  \includegraphics[width=.4\columnwidth]{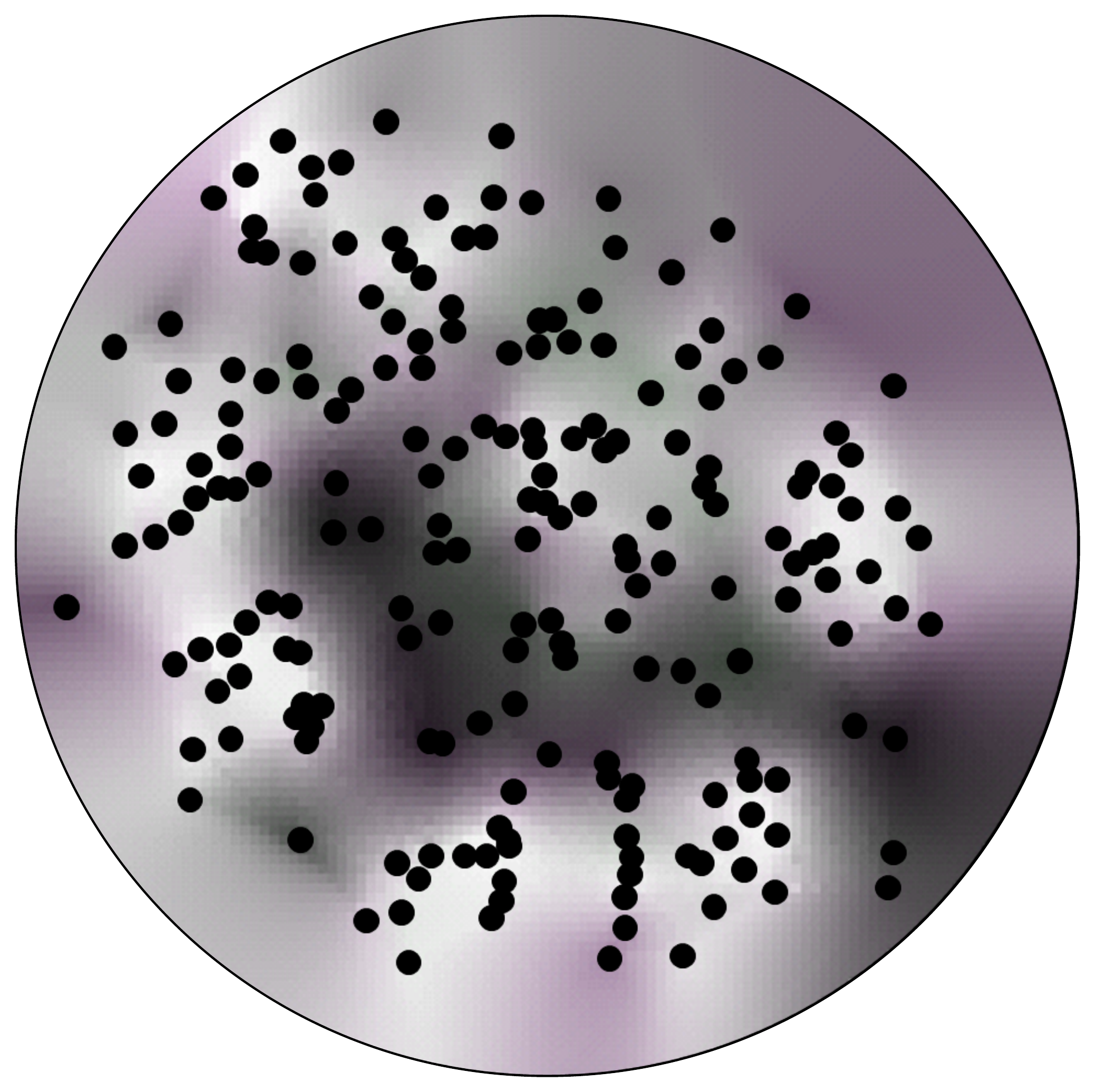}
  \includegraphics[width=.3\columnwidth]{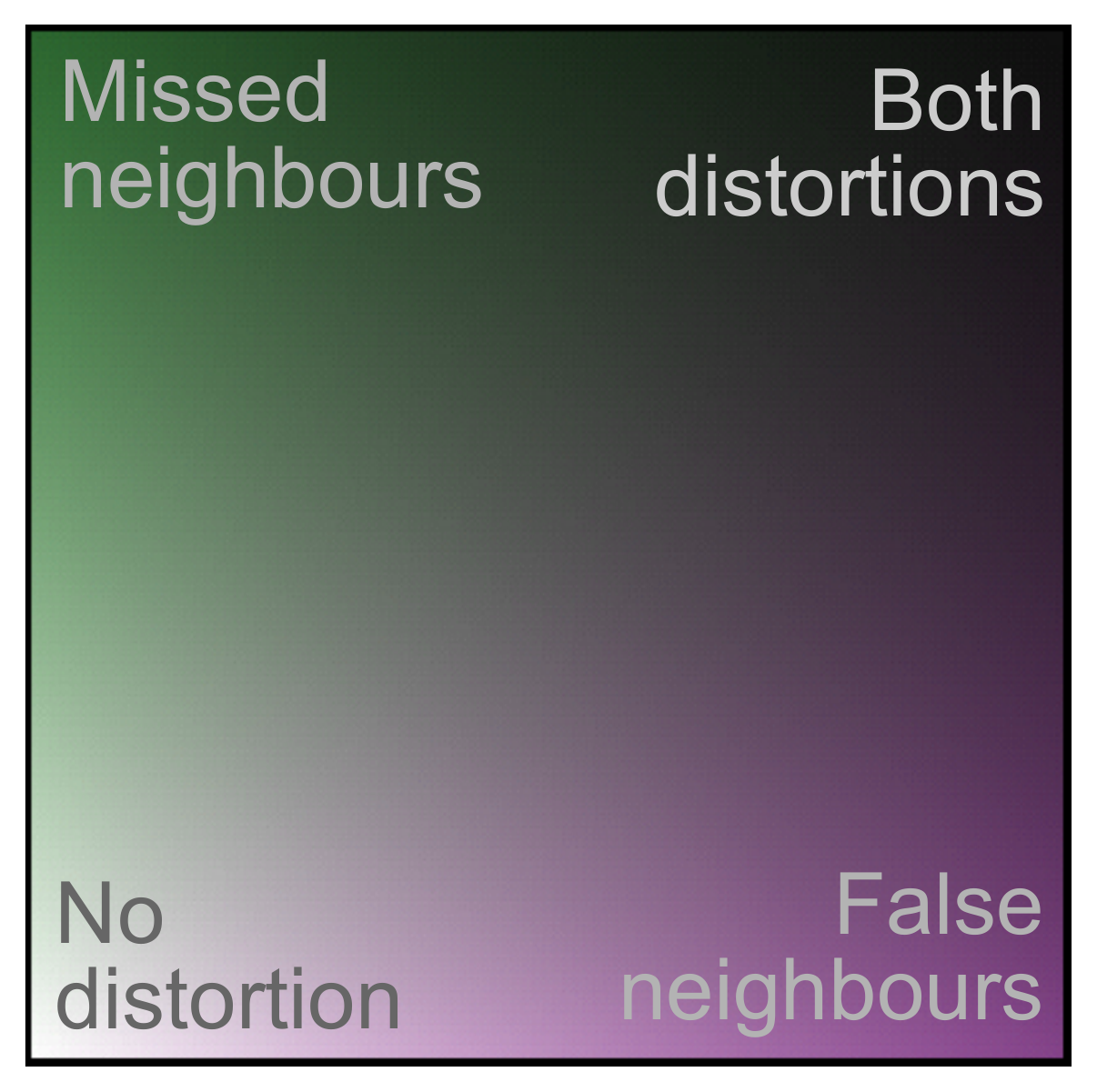}
  \caption{\label{fig:CheckViz}
           \cviz with trustworthiness and continuity weights and $\kappa=10$, for 200 handwritten digits data mapped using \ddhds.
           }
\includegraphics[trim=20 70 30 30,clip,width=1\linewidth]{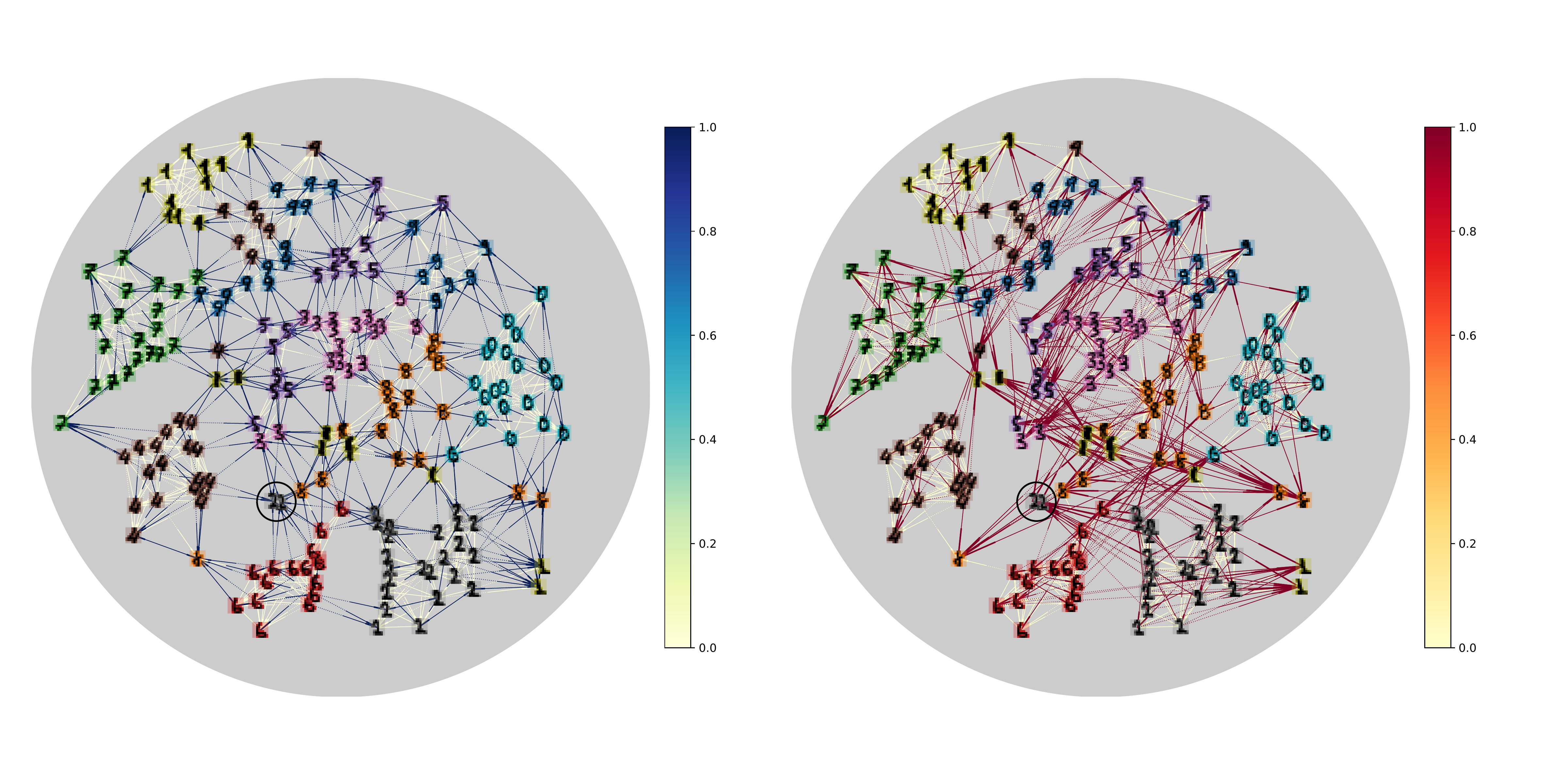}\caption{\label{fig:MingPrecRec}From left to right: \ming retrieval graph $G_{\mathscr{P}}^{\retr}(10)$ and relevance graph $G_{\mathscr{R}}^{\rel}(10)$ with precision and recall weights, for 200 handwritten digits data mapped using \ddhds.}
\includegraphics[trim=20 70 30 30,clip,width=1\linewidth]{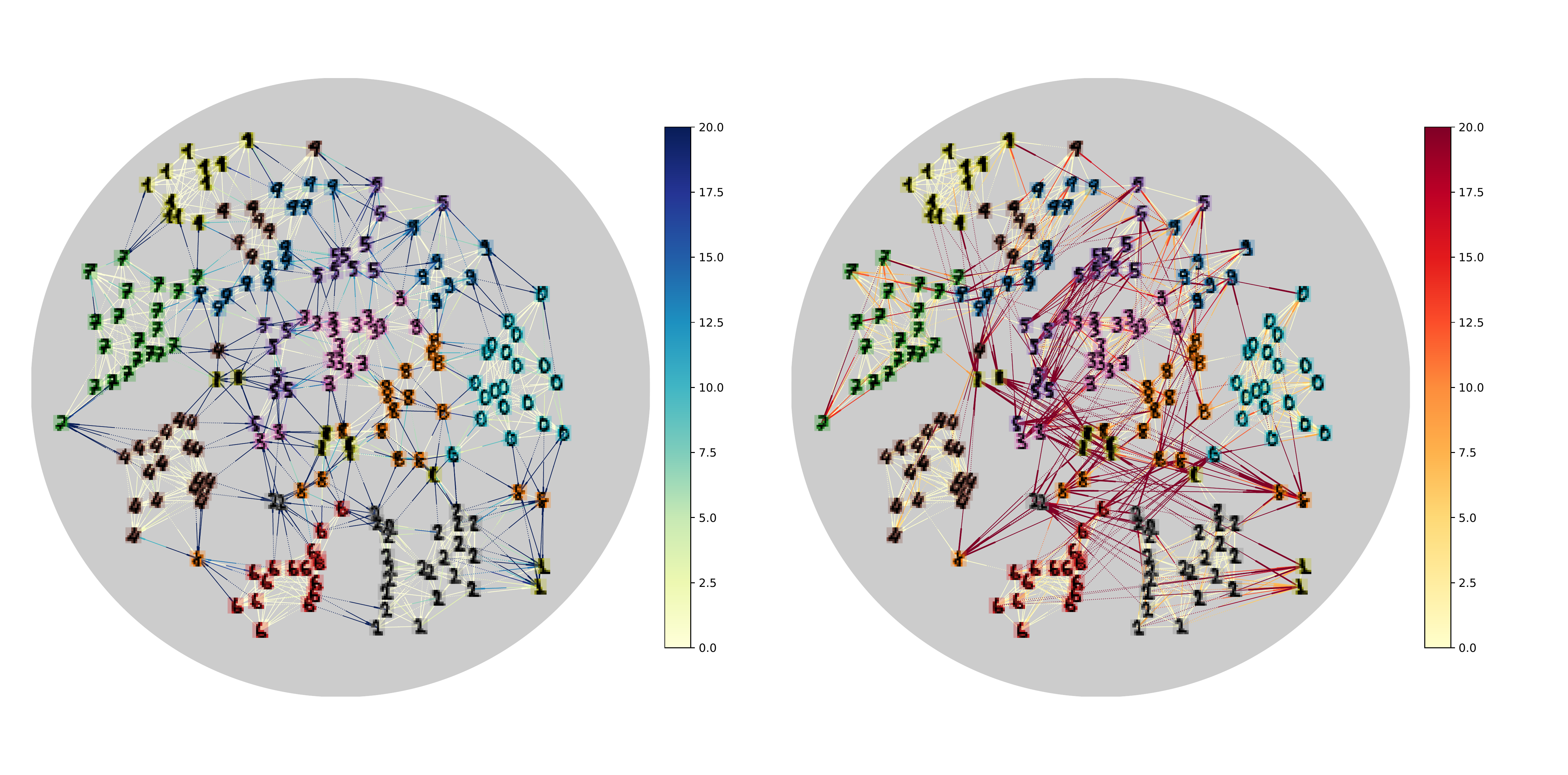} \caption{\label{fig:MingTrustCont}From left to right: \ming retrieval graph $G_{\mathscr{T}}^{\retr}(10)$ and relevance graph $G_{\mathscr{C}}^{\rel}(10)$ with trustworthiness and continuity weights, for 200 handwritten digits data mapped using \ddhds.}
\end{figure}

For the sake of comparison, Figure~\ref{fig:CheckViz} shows \cviz on the background of the same map used for illustrating \ming. 
Results are coherent with \ming about neighbourhood distortions for each point, 
but it fails to inform on which pairwise relations are distorted. 

\subsubsection{Precision and recall weights}
Figure~\ref{fig:MingPrecRec} shows the retrieval and relevance graphs 
for precision and recall weights. 
On the retrieval graph (left panel), the visually discernible group of \ovalbox{2} at the bottom right of the map  
is confirmed by the numerous white edges connecting points together. 
Conversely, this group is linked to the adjacent \ovalbox{6}, \ovalbox{8} and \ovalbox{1} by a majority of blue edges, which shows that they are clearly separated in the data space. 
Note that without the superimposed graphs, a visual clustering could easily lead to wrongly splitting this group of \ovalbox{2} along its central vertical gap. 
A similar reasoning allows to locate the isolated pair of \ovalbox{2} circled on Figure~\ref{fig:MingPrecRec} and lost between \ovalbox{4}, \ovalbox{3}, \ovalbox{8} and \ovalbox{6}. 
Moreover, on the relevance graph, these two \ovalbox{2} are linked to the rest of their class by many red edges, 
indicating that they have been torn apart by the mapping. 
Thus, an analyst could safely conclude with \ming that all the \ovalbox{2} are part of the same cluster, 
and should have been mapped together. 
In the same way, we may identify 
the \ovalbox{1} (top left), \ovalbox{4} (bottom left) and \ovalbox{4} $\cup$ \ovalbox{9} (top) clusters. 
In the mapping, legitimate proximities among distinct classes, supported on Figure~\ref{fig:MingPrecRec} by visual similarity of digits, 
are also presented by the retrieval graph (left panel), for example between the \ovalbox{1} and the \ovalbox{4} $\cup$ \ovalbox{9} clusters, or inside the latter. 
On another note, the relevance graph (right panel) shows points placed between several attracting clusters mapped at distant locations,  
such as the isolated \ovalbox{4} linked by numerous red edges to the \ovalbox{$4 \cup 9$} and \ovalbox{4} clusters.

\begin{figure}[t]
\centering
\includegraphics[trim=10 10 40 10,clip,width=0.85\linewidth]{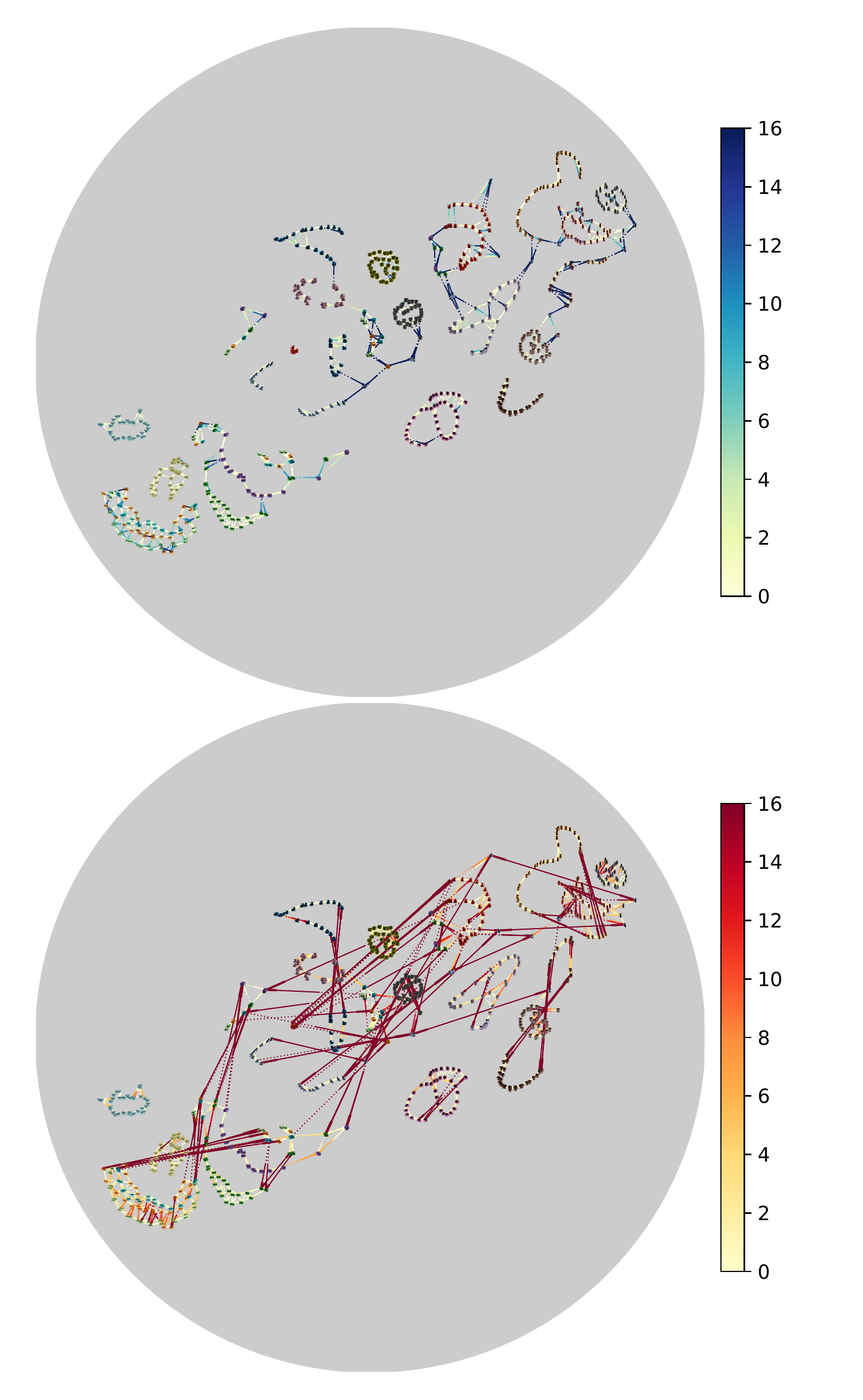}
\caption{\label{fig:coil20_tsne}From top to bottom: \ming graphs $G_{\mathscr{T}}^{\retr}(4)$ and $G_{\mathscr{C}}^{\rel}(4)$ with trustworthiness and continuity weights, for half of the coil-20 dataset mapped with \methodname{t-SNE}.}
\end{figure}

\subsubsection{Trustworthiness and continuity weights}
Since precision and recall weights are binary, slight rank distortions around the cut-off $\kappa$ (\textit{e.g.} tenth neighbour becoming the eleventh) are displayed the same way as severe distortions. 
Thereby, some undesired blue and red edges appear inside clusters on Figure~\ref{fig:MingPrecRec}.  
To allow a softer behaviour, we use trustworthiness and continuity weights (as presented in Table~\ref{table:PairwisePenalizations}). 
In order to increase colour gradient, 
the colour map is saturated at $20$ (namely, all penalizations above this limit are represented with the colour corresponding to maximum penalization), so that rank variations beyond the $30$ neighbourhood ($3\kappa$) are ignored. We may note that the choice of colour saturation is critical in the perception of penalizations, a lower saturation inducing a higher perceived severity for an identical distortion. 
With this weighting, presented by Figure~\ref{fig:MingTrustCont}, 
edges inside clusters turn to lighter shades (green for formerly blue edges, and orange for red), as may be observed by zooming in on the cluster of \ovalbox{$0$}. 
Hence, severely distorted neighbourhood relations tend to stand out, 
while light distortions are still noticeable through a thorough observation.

The coil-20 dataset forms cyclic continuums in the data space, the image of an object changing just slightly from one angular step to another. However, for some classes such as the different models of cars, images are more similar between classes for a specific angle, than they are for the same class with very different rotations. Thus, the mapping cannot perfectly preserve those cycles. 
Figure~\ref{fig:coil20_tsne} presents \ming graphs for the coil-20 dataset (with colour saturation at a penalization of $16$, namely for ranks beyond $5\kappa$). 
On the retrieval graph (top panel), white edges highlight well-represented parts of continuums, but also reliable proximities between different classes. Blue edges show isolated images torn from their group and cycles not closing well (\textit{e.g.} souvenir pigs) or placed too close from another different class (such as the cats). On the relevance graph (bottom panel), red edges show the right way of closing cycles (\textit{e.g.} for pigs or cats), and link isolated images to their main group.

\subsection{Edge bundling}
\begin{figure}[htb]
\centering
\includegraphics[trim=20 70 30 60,clip,width=1\linewidth]{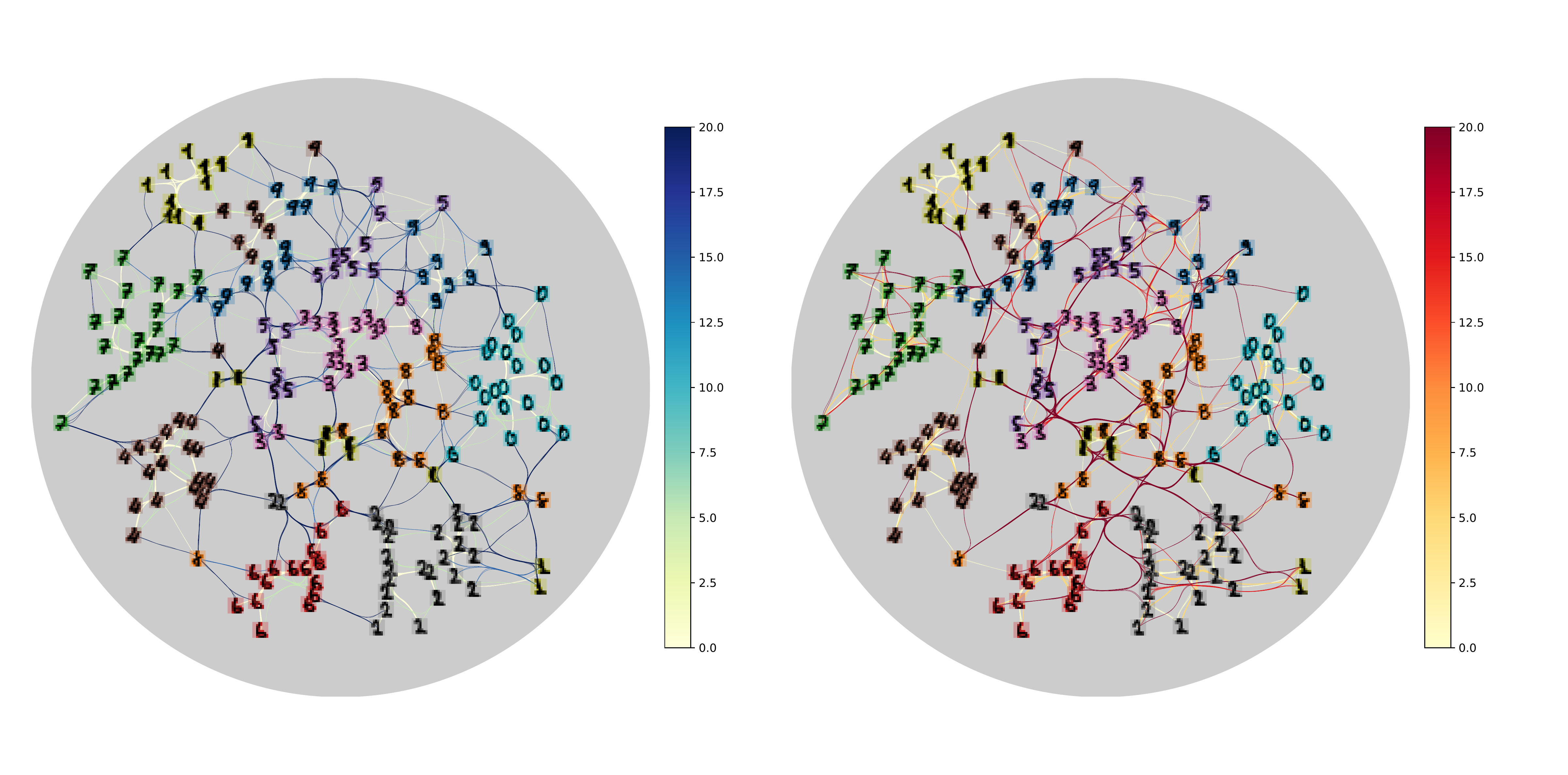}\caption{\label{fig:digits_edge_bundling}Edge bundling with a variant of \methodname{KDEEB} on the trustworthiness and continuity weighted retrieval and relevance graphs for the digits dataset.}
\end{figure}

In order to reduce the visual clutter occasioned by the method, edge bundling was experimented with a variant of the \methodname{KDEEB} algorithm \cite{hurter_graph_2012}, using the previously considered embedding of digits data for the sake of comparison. For bundling together edges whose weights individually convey information, we segregated them in four groups (for each graph) by their associated penalizations value, using the bins $\{0\}$, $]0;10]$, $]10;20]$ and $]20;+\infty[$. We then drew the corresponding bundles using the central colour of each group, resulting in the display shown Figure~\ref{fig:digits_edge_bundling}. 

The obtained visualization is effectively less cluttered, with the structure of frontiers between clusters remaining visible on the retrieval graph. However, the relevance graph suffers from confusion between long red edges, sometimes losing the local information about their source and target. It is for example the case for the edges linking the aforementioned isolated pair of \ovalbox{2} to their main cluster, which are diverted when crossing orthogonal edges starting from the cluster of \ovalbox{6}. Hence, density-based bundling may not be the best suited solution to visual clutter.

Another simple way of reducing the impact of visual clutter could be to interactively draw the edges sorted by increasing penalization. By tuning a penalization threshold, the analyst could first discover the reliable edges, and then add on top of them the more distorted edges, which tend to induce the most clutter in the relevance graph.

\section{Conclusions}
In this paper, we proposed a new method for interpreting mappings 
while accounting for distortions by superimposing two colour-coded graphs. 
These graphs allow to diagnose reliable and distortion-induced neighbourhood relations, 
while quantifying those distortions through a colour scale based on classical map evaluation criteria. 
Therefore, our pairwise display allows to interpret locally the map-wise results obtained for global indicators.
Future work may extend \ming to other pairs of indicators, such as \methodname{MRRE} or \methodname{CCA} and \methodname{Sammon} stresses, or search for ways of reducing visual clutter with the minimal loss of information, especially for missed neighbours edges.

\clearpage

\bibliographystyle{abbrv-doi}

\bibliography{References}
\end{document}